\documentclass[runningheads]{llncs}
\usepackage{graphicx}
\usepackage{amssymb}
\usepackage{amsmath} 

\usepackage{float}
\usepackage{hyperref}

\usepackage{subcaption}

\usepackage{tablefootnote}

\begin{document}
\title{Can deep neural networks learn \\process model structure? \\An assessment framework and analysis}
%
%
\author{Jari Peeperkorn\inst{1} \and
Seppe vanden Broucke\inst{2,1} \and
Jochen De Weerdt\inst{1}}
\authorrunning{Peeperkorn et al.}
\titlerunning{Deep learning process model structure}
%
\institute{Research Center for Information Systems Engineering (LIRIS), KU Leuven, Leuven, Belgium \and
Department of Business Informatics and Operations Management, Ghent University, Ghent, Belgium \\
\email{\{jari.peeperkorn, seppe.vandenbroucke, jochen.deweerdt\}@kuleuven.be}}

\maketitle              
\begin{abstract}
Predictive process monitoring concerns itself with the prediction of ongoing cases in (business) processes. Prediction tasks typically focus on remaining time, outcome, next event or full case suffix prediction. Various methods using machine and deep learning have been proposed for these tasks in recent years. Especially recurrent neural networks (RNNs) such as long short-term memory nets (LSTMs) have gained in popularity. However, no research focuses on whether such neural network-based models can truly learn the structure of underlying process models. For instance, can such neural networks effectively learn parallel behaviour or loops? Therefore, in this work, we propose an evaluation scheme complemented with new fitness, precision, and generalisation metrics, specifically tailored towards measuring the capacity of deep learning models to learn process model structure. We apply this framework to several process models with simple control-flow behaviour, on the task of next-event prediction. Our results show that, even for such simplistic models, careful tuning of overfitting countermeasures is required to allow these models to learn process model structure.

\keywords{Predictive Process Monitoring \and Next Event Prediction \and Recurrent Neural Network \and Generalisation}
\end{abstract}
\section{Introduction}
In the field of process mining, a clear trend can be discerned in terms of a shift from post factum analysis to predictive and even prescriptive modelling. This is for instance clearly reflected in the surge in papers presenting deep learning-based modelling techniques to address analysis tasks including remaining time, outcome, and next event prediction. One potentially problematic issue regarding the application of such deep learning models is the fact that, to the best of our knowledge, no research has focused on investigating whether popular modelling architectures such as LSTM neural networks, can actually ``learn'' process behaviour from a possibly incomplete set of example traces in an event log. 

Accordingly, the main contribution of this work is to propose a framework to assess the capability of deep learning-based next event prediction techniques to truly learn process model structure. The framework consists of a variant-based resampling procedure combined with novel metrics to assess fitness, precision and generalisation of the learned neural network models. In our experimental evaluation, we rely on six relatively (and purposefully) trivial process models that reflect essential control-flow constructs in business processes. By doing so, we can investigate the relation between types of control-flow behaviour (e.g. AND, XOR, and OR split/join, loops, long distance dependencies etc.) and the capacity of deep learning models to truly learn these structural patterns. Our findings indicate that even for such trivial models, and in particular for models with parallel behaviour, rigorous application of overfitting countermeasures is required to have any chance of steering the neural network model towards the goal of truly learning process model structure,  more so when compared with other domains in which such networks have been applied. These findings have important consequences, given that real-life models and event logs are usually orders of magnitude more complex than the models used here. As such, we believe that this paper opens up an important agenda for further investigation. \\

Our paper is organised as follows. First, Section~\ref{Related} discusses some relevant related work. Next, our proposed framework is explained in Section~\ref{framework}. In Section~\ref{Experiments}, the carefully selected artificial process models are proposed, before discussing the hyperparameter grid search and presenting the experimental results. What follows is a brief discussion of the results and their implication (Section~\ref{Discussion}). The paper is concluded in Section~\ref{Conclusion}, which also provides an outlook towards  future research. The synthetic data, results and trained models used and presented in this paper are available  online\footnote{\url{https://github.com/jaripeeperkorn/GeneralizationPPM}}.

\section{Related work}
\label{Related}

In recent years, within the field of Predictive Process Monitoring (PPM), a lot of attention has been attracted by deep learning-based solutions, most frequently Recurrent Neural Networks (RNN)~\cite{Bukhsh_2021,Camargo_2019,Evermann_2017,Lin_2019,Tax_2017,Taymouri_2020}. Given the scope of this paper, we limit this section to a selection of PPM works addressing the next event prediction problem, i.e. given a prefix of activities, produce a probability vector corresponding with the likelihood of each respective activity occurring as the next one. In their pioneering work, Tax et al.~\cite{Tax_2017} propose to use a Long Short-Term Memory network (LSTM) to predict next events and corresponding timestamps, thereby relying on one-hot encoding of the activity labels as input for the LSTM, together with their timestamps. Moreover, Evermann et al.~\cite{Evermann_2017} also propose the use of LSTM networks, specifically to predict full case suffixes rather than the next event only. However, they reduce the input dimension of the event labels by using vector embeddings, and include attributes such as resources. Camargo et al.~\cite{Camargo_2019} use separately trained embeddings of categorical variables and timestamps in order to predict both next events as well as future timestamps. Moreover, Lin et al.~\cite{Lin_2019} use an LSTM encoder-decoder and a modulator structure to predict next events and suffixes, using both control-flow information as well as other event attributes. Recently Taymouri et al.~\cite{Taymouri_2020} proposed a Generative Adversarial Networks approach to the problem of next event, suffix and timestamp prediction showing promising results. In Bukhsh~\cite{Bukhsh_2021} a transformer network approach is proposed to the problem of next event prediction. Transformer networks have recently been used to beat several benchmarks in other fields like Natural Language Processing~\cite{attentionisallyouneed}. 

Despite the drastic increase in research attention, no studies have investigated whether RNN-based architectures can actually learn process model structure. As such, it is unclear whether the generalisation that is expected from plain process discovery techniques is realised by neural network models. It has been shown that RNNs are universal approximators~\cite{UniversalApproximators}. And while Siegelmann and Sontag~\cite{siegelmann1995computational} showed in 1995 already that RNNs are Turing complete, i.e. for any given computable function there exists a finite RNN to compute it, there is still much work on understanding what makes functions difficult to learn, let alone under the constraint of data incompleteness as is the case with business process data sets~\cite{reallife}. That is, whilst it might be clear that RNNs are capable to fit the given training data in the form of process cases, the central question we investigate here is whether such models can be constructed so that they are also able to generalise towards making good predictions for new unseen control-flow behaviour, which is highly likely to occur once a process starts to exhibit even a limited amount of complexity. In other words: do these models memorise the training data or truly learn the process structure?

\section{A framework for assessing the generalisation capacity of RNNs}
\label{framework}

With the goal of this paper in mind, we set out on developing a framework that is capable to assess to what extent RNN-based architectures are capable to learn process model structure. For an introduction on how RNNs work and how they can be used in predictive process monitoring, the interested reader is referred to~\cite{Evermann_2017}. This framework relies on a specific resampling procedure combined with a set of new metrics to quantify recall, precision and generalisation. Recall that we restrict ourselves to next event prediction models, however, an extension to suffix prediction is trivial. Moreover, given that many remaining time or outcome prediction models also rely on incorporating control-flow information, it can be expected that these models too would benefit from proper generalisation, if at all possible. 

\subsection{The resampling procedure}

\begin{figure*}[htbp]
\centerline{\includegraphics[width=.95\linewidth]{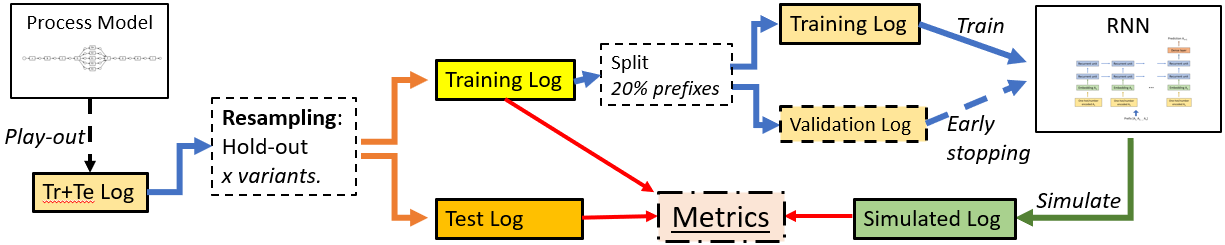}}
\caption{Overview of the setup.}
\label{fig:overview}
\vspace{-6mm}
\end{figure*}

A schematic overview of the assessment framework is shown in Figure~\ref{fig:overview}. We start from a (simple) process model and play-out the model to obtain a corresponding event log. We assume that we work in a setting where the number of variants (i.e. distinct traces in terms of the events and their order), is bounded. Therefore, in case of loops, we assume a maximum number of times a certain marking can be visited. Once the event log is obtained, we determine all of its unique control-flow variants. Next, a resampling procedure at the level of variants is performed to construct training and test sets. For instance, one can decide to simply retain all cases pertaining to one single variant in the test log, resulting in a ``leave-one-variant-out cross-validation'' (LOVOCV).

Thus, starting from the complete event log, which is referred to as the \textit{Train+Test log (Tr+Te)}, we single out all cases pertaining to one or more variants to form the \textit{Test log (Te)}. The remaining variants form the \textit{Training log (Tr)}, which is split into all possible prefixes to train the model. From this set of prefixes, a \textit{Validation log (Val)} is created, mainly to allow the training procedure of the LSTMs to use early stopping. For now, we simply perform a random selection of 20\% of the \textit{Training log}'s prefixes. The \textit{Training} and \textit{Validation log}'s prefixes are then used to train a model for next event prediction using the observed next events of every prefix as the target. Accordingly, the model is trained to predict for every prefix what the subsequent event's activity label will be. As such, the model outputs a probability for each of the different activities in the activity vocabulary. Once trained, we use the RNN model to simulate a \textit{Simulated log (Sim)} as follows. We start by presenting the RNN with a prefix only containing the beginning of sequence token (BOS). As output, the model returns a probability for each of the possible activity labels to be the next event. Using these probabilities, we sample a possible next event, to be appended to the existing prefix. Subsequently the prefix is used to sample a next event in the same way. We do this until we reach the end of sequence token (EOS) or a certain predetermined maximum size is reached. LSTMs have been used as generative models similarly in~\cite{Camargo_2019}. The idea is that, by simulating an event log from the RNN, we should ideally obtain an event log that is behaviourally highly similar to the event log that we started from. That is, we expect that, even when leaving out one single variant, the RNN should be able to (1) generalise this variant from the observed variants in the Training log, (2) avoid creating variants that were not observed in the original event log (Train+Test), and (3) contain all the variants present in the Training log.

\subsection{Metrics}

Accordingly, we define novel recall, precision and generalisation metrics that can quantify these three criteria. 
Based on the \textit{Training} (including \textit{Validation}), \textit{Test} and \textit{Simulated} logs, we define the following metrics: 

\begin{equation}
    \textit{Fitness} = \sum_{v\in\textit{Var(Tr)}} \frac{\textit{Min}\left(\textit{Occ(}v\textit{,Sim)}\textit{, Occ(}v\textit{,Tr)}\right)}{|\textit{Tr}|}
\end{equation}
\begin{equation}
    \textit{Precision} = \sum_{v\in\textit{Var(Sim)}} \frac{\textit{Min}\left(\textit{Occ(}v\textit{,Sim)}\textit{, Occ(}v\textit{,Tr+Te)}\right)}{|\textit{Sim}|}
\end{equation}
\begin{equation}
    \textit{Generalisation} = \sum_{v\in\textit{Var(Te)}} \frac{\textit{Min}\left(\text{Occ(}v\textit{,Sim)}\textit{, Occ(}v\textit{,Te)}\right)}{|\textit{Te}|}
\end{equation}
with $|L|$ denoting the number of traces in an event log $L$, $\textit{Var(L)}$ denoting the set of variants of an event log $L$ and $\textit{Occ(v,L)}$ a function denoting the frequency or multiplicity of a variant $v$ in an event log $L$.

Each of these metrics outputs a value between 0 and 1. Beware that these metrics make use of nominal counts, so they only make sense when the original \textit{Train+Test log} and the \textit{Simulated log} contain the same amount of traces. If not, the metrics will have to be corrected. First of all, the fitness metric measures to what extent all of the variants present in the \textit{Training log} are also present in the \textit{Simulated log}. This is because we want the RNN to learn and replicate all of the behaviour found in the \textit{Training log}. Moreover, we expect that the frequency of each variant in \textit{Simulated log} is, more or less, equal to the frequency of observation of that variant in the \textit{Training log}. Therefore, the fitness measure will punish if a certain variant is under-represented in the \textit{Simulated log}. 
Secondly, the precision metric measures whether the RNN allows for too much behaviour, i.e. traces that have not been seen in \textit{Train+Test log}. Moreover if certain correct variants are over-represented in the \textit{Simulated log} the precision will also decrease. Finally, the generalisation metric quantifies to which extent the RNN is able to generalise, i.e. whether it is able to learn and reproduce correct but unseen behaviour. Therefore, the metric measures whether the frequency of occurrence of the unseen variant(s) in the \textit{Test log} is actually reproduced to the same level in the \textit{Simulated log}. 

\section{Experimental evaluation}
\label{Experiments}

With the introduction of our assessment framework, consisting of a variant-level resampling procedure combined with these three metrics, we can now devise an experimental setup to evaluate the generalisation capacity of RNNs. While it is theoretically possible to perform such a assessment using complex artificial and even real-life logs and models, we opt to focus on simple models, as these provide a \textit{sine qua non} condition in terms of investigating whether such models can deal with the aforementioned control-flow patterns at all.

\subsection{Process models}
\label{Models}

\begin{figure*}[h!]
\captionsetup[subfigure]{labelformat=empty}

\begin{subfigure}{0.49\textwidth}
  \includegraphics[width=\linewidth]{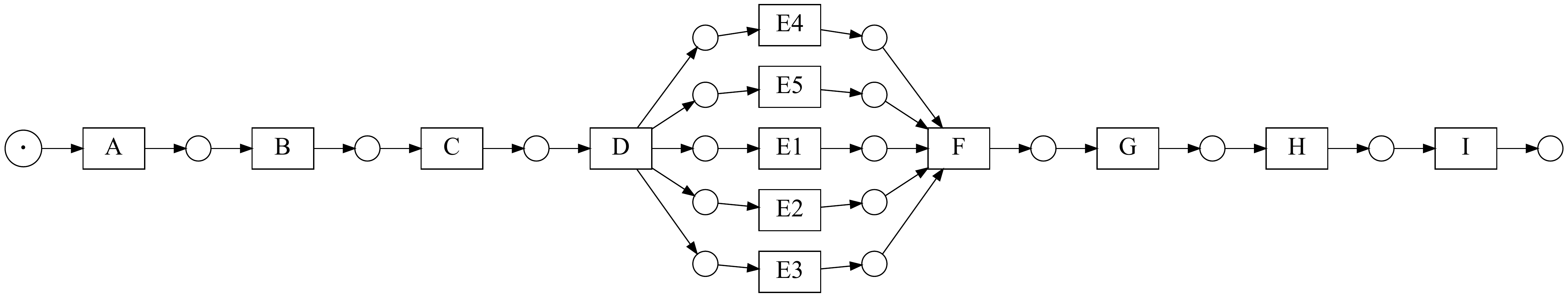}
\caption*{Model 1: Parallel Model with 120 variants.}
\end{subfigure}
\hfill
\begin{subfigure}{0.49\textwidth}
  \includegraphics[width=\textwidth]{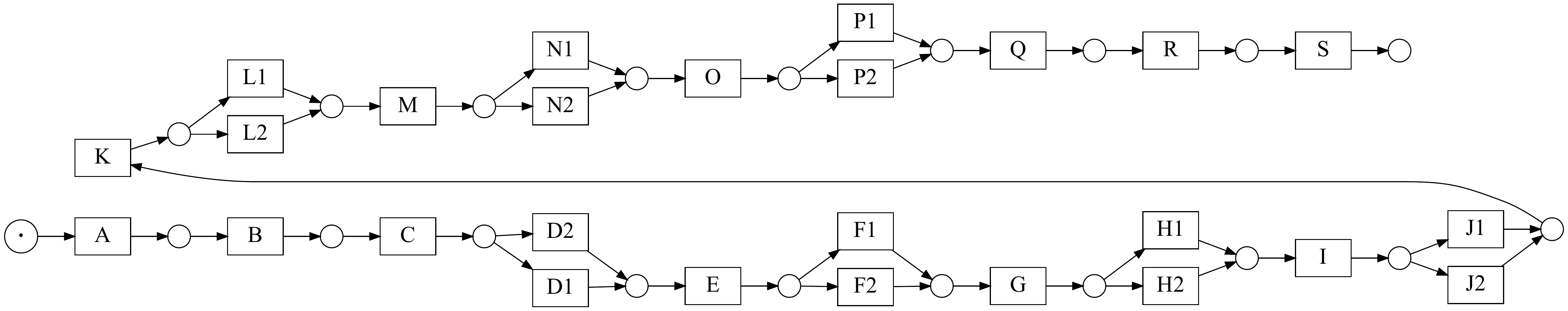}
    \caption{Model 2: Model with multiple XOR splits, 128 variants.}
\end{subfigure}
\hfill
\begin{subfigure}{0.49\textwidth}
  \includegraphics[width=\textwidth]{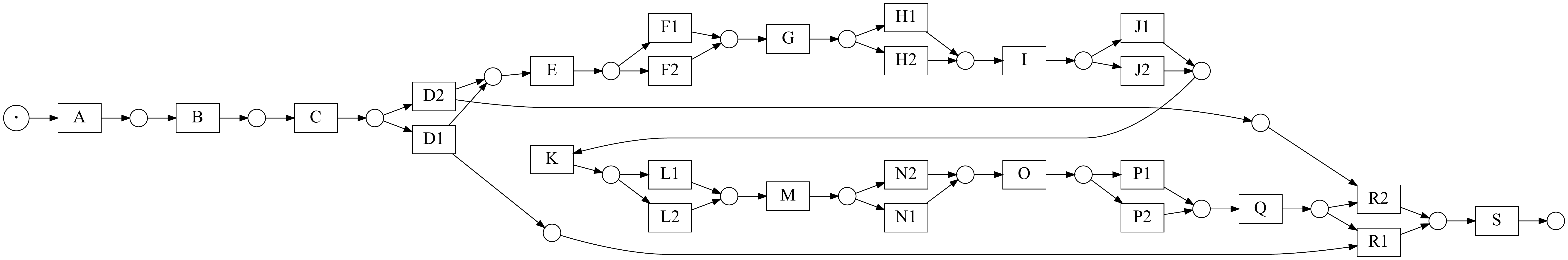}
\caption{Model 3: Model with multiple XOR splits and a long term dependency, 128 variants.}
\end{subfigure}
\hfill
\begin{subfigure}{0.49\textwidth}
  \includegraphics[width=\textwidth]{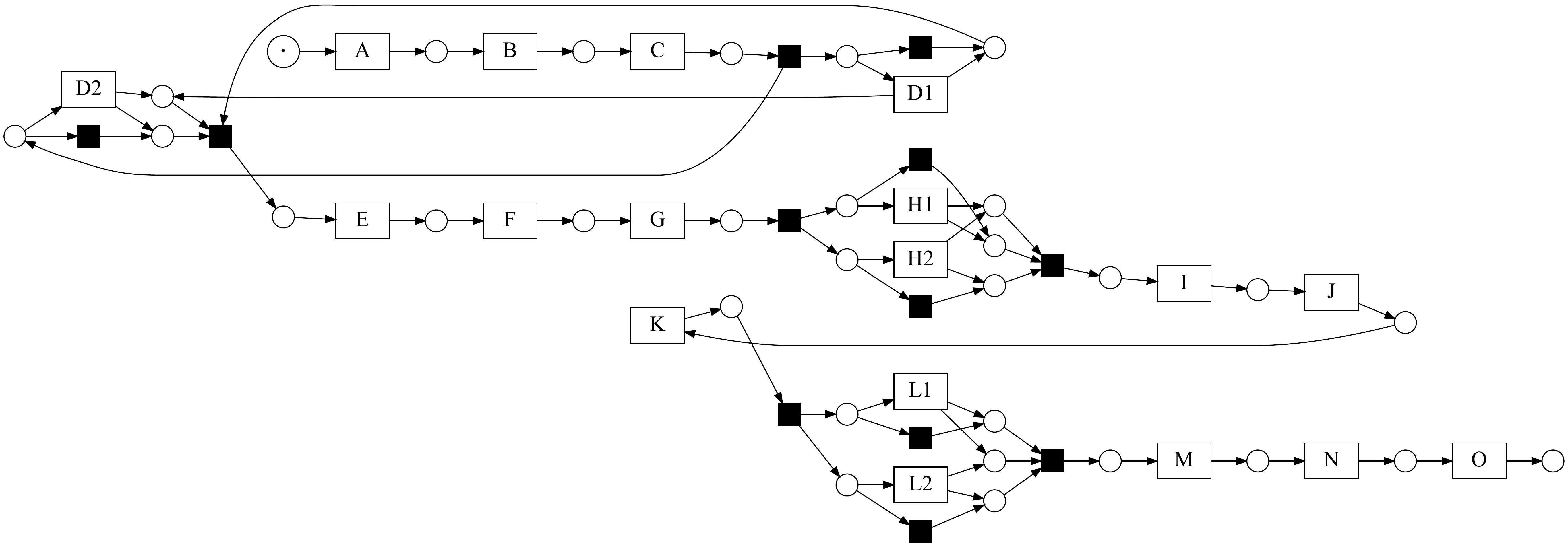}
  \caption{Model 4: Model with multiple IOR splits, 64 variants.}
\end{subfigure}
\hfill
\begin{subfigure}{0.49\textwidth}
  \includegraphics[width=\textwidth]{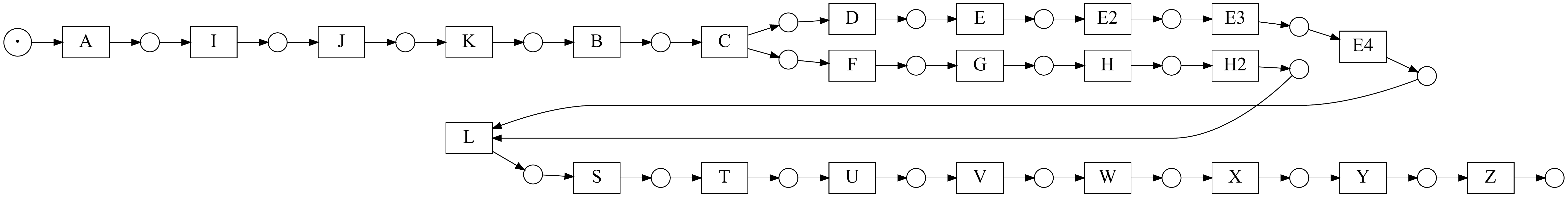}
 \caption{Model 5: Model with one big parallel split, 126 variants.}
\end{subfigure}
\hfill
\begin{subfigure}{0.49\textwidth}
\includegraphics[width=\textwidth]{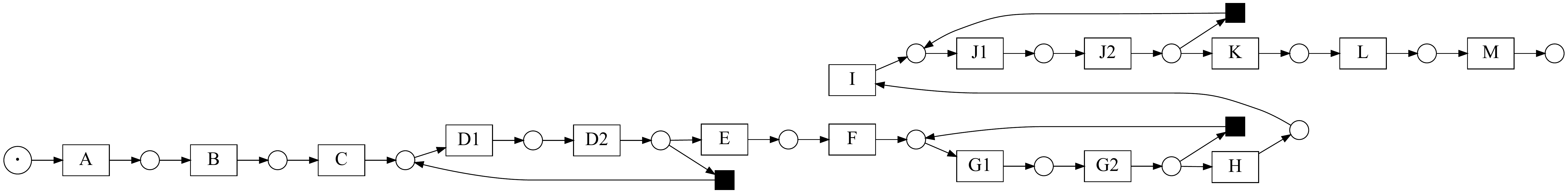}
\caption{Model 6: Model with 3 different size 2 loops.}
\end{subfigure}
\hfill
\vspace{2mm}
 \caption{Process models used in the experimental evaluation.}
\label{fig:models}
\vspace{-8mm}
\end{figure*}

Hence, we generated artificial process models that represent main modelling constructs. In order to obtain the full event log (i.e. the \textit{Train+Test log}), we  use the play-out functionality of the Python Process Mining library PM4PY~\cite{pm4py}. The models are depicted as Petri nets in Figure~\ref{fig:models}. Model 1 is a simple linear model with a  parallel gateway consisting of five parallel branches containing each one single activity. This process has 120 (equally likely) control-flow variants. In Model 2, a process model with 128 (equally likely) control-flow variants is created by sequencing seven exclusive OR (XOR) splits. Similarly, Model 3 consists out of eight XOR splits, but with a long-term dependency added in. Model 4 consists of three inclusive OR (IOR) splits, where at least one, but possibly both of the two activities have to occur. This leads to in total 64 control-flow variants. Model 5 shows a process which has two parallel paths, consisting of five activities each, leading to 126 control-flow variants with varying likelihood. Finally, Model 6 shows a process with three different possible loops (containing two activities each). The amount of possible control-flow variants is technically unlimited, so that we restrict each marking to be visited a maximum of three times, we keep 27 different variants. 

\subsection{Hyperparameter search}
\label{Hyper}

The generalisation capacity of RNNs strongly depends on tuning its hyperparameters. This is an essential part of the training procedure. An overview of the investigated hyperparameters can be found in Table~\ref{tab:grid}. A maximum prefix length of size $10$ was used (with longer prefixes left-truncated), unless explicitly mentioned otherwise.  The model's weights are optimised using the Adam~\cite{Adam} optimiser with a mini-batch size of $128$ prefixes, using a starting learning rate of $0.005$. The learning rate is decreased when the accuracy on the validation set has not decreased for over 10 epochs and the training is stopped (early stopping) when the accuracy has not increased for over 30 epochs (or when a maximum of 600 epochs is reached). The loss function used is categorical crossentropy. For clarity, the RNN is trained optimising accuracy in a ``classical'' sense, i.e. whether the activity predicted by the model to be the most probable activity is actually correct. As mentioned earlier, the usage of an embedding layer is the first binary hyperparameter. Where applicable, the dimension of the embedding was set to $\lceil\sqrt[\leftroot{-2}\uproot{2}4] {\text{Act. Voc. Size}}\rceil$, as was done in~\cite{Camargo_2019} (note that the embeddings were pretrained independently from the RNN in that work, though we used this value as a starting point in this investigation). The number of stacked LSTM layers is varied between one and two, with the layers' hidden dimension size set to 16, 32 or 64 units. Furthermore, we experiment with different values for overfitting countermeasures like regularisation and dropout. We try five different values for \textit{L1} (\textit{Lasso}) and \textit{L2} (\textit{Ridge}) regularisation~\cite{regularization}, which adds a small penalty to the model based on the absolute or squared value of the weights respectively. In order to limit the scope of our grid we change $L1$ and $L2$ together. Lastly we also added the dropout hyperparameter~\cite{dropout}, in which, per epoch, a fraction of the nodes is selected to be ignored during training, reducing the likelihood of overfitting to the training data. In this paper we experiment with multiple dropout values (including no dropout). There are multiple ways to introduce dropout when using RNNs related to the internal structure of such units; in this work we chose to add dropout to the output of the LSTM layer (note that dropout on the inputs would lead to removing certain steps of a prefix which is undesirable in our setting).  We omitted the use of batch normalisation due to its limited effectiveness when applied to RNNs~\cite{Cooijmans_2016}. In future work, however, we could explore the effectiveness of \textit{Recurrent Batch Normalisation}~\cite{Cooijmans_2016} and \textit{Layer Normalisation}~\cite{Ba_2016}. The neural network implementation was created using the Python library Keras\footnote{\url{https://keras.io}}.

\begin{table}[ht]
    \centering
    \begin{tabular}{c|c}
        \textbf{Hyperparameter} & \textbf{Values} \\ \hline
        Embedding Layer & Yes, No \\
        Number of LSTM Layers & 1, 2 \\
        LSTM Layer Size & 16, 32, 64 \\
        L1 and L2 & 0.0, 0.00001, 0.0001, 0.001, 0.01 \\
        Dropout & 0.0, 0.2, 0.4
    \end{tabular}
    \caption{The hyperparameter values used in the grid search. \label{tab:grid}}
    \vspace{-4mm}
\end{table}

This led to 180 different hyperparameter configurations to iterate over. Based on some preliminary exploration, it was noticed that an RNN, without explicit overfitting countermeasures like regularisation and dropout, struggled to generalise for the process in Model 1. In order to contain computational time, and because the goal is to obtain a RNN that is able to generalise different types of behaviour at the same time, we opted to only perform one grid search on the event log obtained from this process model. The best hyperparameter settings for Model 1 were subsequently used and applied to all other models. 
For obtaining optimal hyperparamaters using Model 1, we conducted a tailored leave-one-variant-out cross-validation (LOVOCV) procedure as introduced above. More specifically, we generated a \textit{Train+Test log} consisting of 12.000 traces for each model. In each LOVOCV iteration, we singled out all cases pertaining to one single variant into the \textit{Test log}. For every hyperparameter combination, we performed the tailored LOVOCV eight times, each time with a different variant in the test set. In each iteration, we obtained a \textit{Simulated log} of equal size of the original \textit{Train+Test log}. Based on these eight LOVOCV iterations, we calculated the three different metrics defined above, and, for each setting, took the average over the eight iterations. Note that we use the \textit{Test log} for this hyperparameter tuning, rather than the \textit{(cross) validation log} as is usual. Since we are not trying to compare the predictive quality of different approaches as such, this is justified. We choose to continue working with the setting showing the highest average score over all three metrics, i.e., no embedding layer, one LSTM layer of hidden size 32, an \textit{L1} and \textit{L2} of $0.001$ and a \textit{dropout} of $0.4$. Because this model was only trained on the data of one simple process model, and the differences between certain settings were slim, we however do not want to claim this setting is ideal for each predictive process monitoring problem. However in the context of this investigation rather than optimisation experiment, we continue with this setting in the rest of the paper. Not shown here, but apparent from the hyperparameter search was: (i) for all three metrics a regularisation value of $0.01$ resulted in low scores, (ii) no or limited application of overfitting countermeasures leads to high fitness and precision (as expected) but weak generalisation scores. We like to stress that it is therefore of the utmost importance to tune the hyperparameters correctly when training RNNs.

\subsection{Results}
\label{Results}

Subsequently, we then repeat the experiment using the settled hyperparameter configuration for all models. For each of these, we applied a LOVOCV setup again, working with  \textit{Test logs} containing a single variant. As for some models the frequency of the variants is not evenly distributed and to obtain more robust results, we now conducted an exhaustive LOVOCV, i.e. the procedure is repeated as many times as the number of variants in the event log, so that every variant is used once to form the \textit{Test log}. In the case of Model 6 (loops), we restricted the analysis to variants for which the loop is taken a maximum of three times (27 variants). This was however not restricted neither in the play-out itself, nor in the simulation with the trained LSTM, leading to more different variants in these logs. In the experiment for Model 3 (including the long-term dependency) the prefix length of the input of the RNN was set to the maximum trace length minus one instead of the default of $10$ because it needed to be long enough in order to have a chance of dealing with the long-term dependency. The average values over all variants for each of the three metrics can be found on the left side of Table~\ref{tab:Results}. The error intervals are calculated by taking the standard deviation over all the metric values. Various interesting observations can be made from the results. First, it can be noticed that models with parallel behaviour (Model 1, 4 and 5) are problematic for the LSTMs. One can observe some level of generalisation, but given the extremely lenient LOVOCV-setup, it is remarkable that the generalisation scores go well below 0.80. On the other hand, the LSTMs show to be much more robust when dealing with XOR-splits (Model 2 and 3) and loops (Model 6). Also the long-term dependency in Model 3 seems to be handled well by the LSTM models. The standard deviations are significantly higher for the generalisation scores as this metric seems to be more prone to fluctuations. This is most likely due to to the changing \textit{Test log}, though should be further investigated.

Up until this point we have used the most trivial setting, i.e. the leave-one-variant-out \textit{Test log}. However, one might expect that LSMTs should be able to cope with larger fractions of unseen behaviour. Therefore, this final evaluation part addresses the use of larger test sets, in particular, leaving out 20\% of the control-flow variants from the \textit{Training log}. We repeated this three times, with each time 20\% randomly selected variants. The average of the metric values for each of these experiments can be found on the right side of Table~\ref{tab:Results}. The error intervals are calculated by taking the standard deviation over the metric values for the three different experiments. When comparing this with the results from the LOVOCV experiment we can see that for some models, precision seems the decrease a bit. This might suggest that because the model has less behaviour to learn the correct process model structure from, it fares worse, allowing for some extra incorrect behaviour. Fitness seems not to be affected. Generalisation also results in lower values when having a more diverse \textit{Test log}. This is especially apparent in the process models which already yielded low generalisation scores above, like Model 4 and 5. Again the standard deviation is higher when calculating the generalisation, because it is highly dependent on which variants exactly were in the \textit{Test log}, and of the apparently more fluctuation-prone behaviour of this metric. 

\begin{table}[ht]
    \centering
    \setlength\tabcolsep{3pt}
    \begin{tabular}{|c|c|c|c||c|c|c|}
        \hline
        & \multicolumn{3}{c||}{LOVOCV} & \multicolumn{3}{c|}{Leave 20\% out} \\ \hline   
        \textbf{Model} & \textbf{Prec.} & \textbf{Fit.}  &\textbf{Gen.}  & \textbf{Prec.} & \textbf{Fit.}  &\textbf{Gen.} \\ \hline
        Model 1 & $0.94\pm0.00$ & $0.94\pm0.00$ & $0.79\pm0.12$ & $0.89\pm0.01$ & $0.93\pm0.00$ & $0.72\pm0.04$ \\ \hline
        Model 2 & $0.94\pm0.00$ & $0.94\pm0.00$ & $0.92\pm0.09$ & $0.92\pm0.01$ & $0.93\pm0.01$ & $0.89\pm0.04$ \\ \hline
        Model 3 & $0.94\pm0.00$ & $0.94\pm0.00$ & $0.91\pm0.10$ & $0.92\pm0.02$ & $0.93\pm0.01$ & $0.85\pm0.05$ \\ \hline
        Model 4 & $0.95\pm0.01$ & $0.95\pm0.00$ & $0.75\pm0.13$ & $0.87\pm0.03$ & $0.94\pm0.01$ & $0.61\pm0.14$ \\ \hline
        Model 5 & $0.92\pm0.01$ & $0.92\pm0.01$ & $0.68\pm0.21$ & $0.84\pm0.01$ & $0.94\pm0.01$ & $0.47\pm0.05$ \\ \hline
        Model 6 & $0.93\pm0.01$ & $0.93\pm0.01$ & $0.92\pm0.11$ & $0.92\pm0.01$ & $0.93\pm0.00$ & $0.83\pm0.12$ \\ \hline
    \end{tabular}
    \caption{The results on the different process models, averaged over all leave-one-variant-out experiments with every different control flow variant. And the results on the different process models when taking a \textit{Test log} consisting of 20\% of the control flow variants. Average over three different randomly selected \textit{Test logs}. \label{tab:Results}}
    \vspace{-8mm}
\end{table}

\section{Discussion}
\label{Discussion}

From the hyperparameter tuning, it appears that overfitting countermeasures like regularisation and dropout are important parameters for constructing predictive RNNs. However, precise tuning is crucial, and we therefore urge other researchers to perform a good hyperparameter experiment when training RNNs on predictive process monitoring tasks. This may require the inclusion of resampling methods as introduced here, as opposed to only holding out randomly selected prefixes. Please observe that the ``best'' hyperparameters were selected on the test observations for Model 1. It can be expected that standard tuning on a random validation set and thereby only optimising accuracy of next events (and not the metrics presented here), is unlikely to result in the same outcome.

LSTMs seem to be less suited for generalising process models with parallel behaviour. When the degree of incompleteness is increased, i.e. the amount of variants not seen by the RNN during training is expanded, LSTMs seem to struggle more. The generalisation, as well as the precision, decreased when increasing the amount of variants in the test set, when compared with the LOVOCV experiments. 

It could be interesting to investigate to what extent the generalisation and overfitting problems could affect predictions of real-life processes, considerably more complex than the artificially created data discussed in this work. Extra overfitting measures may need to be included as well in the future. One option could be found in using a similar resampling method as used here  in order to construct the \textit{Test log}, or a hybrid approach, to create the \textit{Validation log}. Another course of action would be to alter the loss function used to train the RNN. Important to check further is whether low generalisation scores of merely memorising and overfitting models also lead to less accurate next event predictions. We still assume proper generalisation would be beneficial in predictive task, especially with increasing complexity, and theoretically the power of deep learning models, actually lies in their generalisation capability~\cite{Kawaguchi_2020}. Moreover if overfitting would not be an issue, more explainable statistical models could easily be found as well, as opposed to the black-box LSTMS, with comparable accuracy, and preference should be given to these more explainable models.

\section{Conclusion and Future Work}
\label{Conclusion}

This paper addressed the so far largely uninvestigated problem of neural networks' capability to learn the behaviour of the underlying process behind an event log. By introducing a new framework that combines a variant-level resampling scheme with three novel metrics, we were able to investigate to what extent LSTMs trained to predict the next event of a process execution are general, fit and precise. By applying this framework on several simple process models, it was shown that LSTMs can only generalise parallel behaviour to a certain extent. Even in the most lenient setting, the LOVOCV, the generalisation metric does not return values close to optimal. When increasing the amount of variants in the \textit{Test log}, unseen by the LSTM during training, generalisation decreases further, as well as precision. This paper opens up the door for future work in predictive process monitoring to include explicit generalisation checks as well, in model selection, hyperparameter tuning and testing. 

The experiments in this work were limited to only use control-flow behavior. However, since the generalization behavior investigated in this paper is only control-flow like, this choice is justified. Nevertheless, it might be interesting to investigate the effects of adding more dimensions (like timestamps and resource) to the predictive models. In future work, more synthetic logs could be investigated deepening the relation between process model behaviour and RNN generalisation. A more rigorous theoretical elaboration, similar to~\cite{Tu2020Understanding}, might provide some interesting insights as well. Furthermore, it should be investigated whether new recommendations can be proposed regarding how to optimally sample training and validation sets. Also, it would be interesting to apply these metrics on models trained on real-life event logs, as opposed to only synthetic data. In this paper it was opted to first work with synthetic data since it would allow us to test different types of behaviour independently. However real-life logs are in general more complex, and in this way might present other difficulties. When doing this, it might be useful to expand the hyperparameter grid and include extra parameters such as Layer Normalization~\cite{Ba_2016} and Recurrent Batch Normalization~\cite{Cooijmans_2016}, and to ameliorate the overfitting measures applied in this work. Other future work can address a comparison of results presented here with the findings in recently proposed work by~\cite{Weinzierl_2020}, and similarly test multiple encoding techniques like hash encoding. Moreover, an additional investigation of the attention mechanism seems worthwhile~\cite{Bukhsh_2021}. In addition, a similar experiment could be applied to alternative architectures including Convolutional Neural Networks, Transformer Networks, and GAN-style approaches. This work was also limited to only include next-event prediction models, thus it might be useful to expand the metric definitions to also be able to evaluate full suffix prediction (or even remaining time prediction). Finally, the application of neural networks to formal languages has been investigated in other domains, e.g. in~\cite{Michalenko_2019}, which could lead to more fundamental research on grammar learning capabilities of RNNs.
%
%
%
\bibliographystyle{splncs04}
\bibliography{ref}

\end{document}